# *Static Free Space Detection with Laser Scanner using Occupancy Grid Maps*

Hesham M. Eraqi[a], Jens Honer[b], Sebastian Zuther[b]
[a] Valeo, Cairo, Egypt
[b] Valeo Schalter Und Sensoren, Bietigheim, Germany
hesham.eraqi@valeo.com, jens.honer@valeo.com, sebastian.zuther@valeo.com

***Abstract*— Drivable free space information is vital for autonomous vehicles that have to plan evasive maneuvers in real-time. In this paper, we present a new efficient method for environmental free space detection with laser scanner based on 2D occupancy grid maps (OGM) to be used for Advanced Driving Assistance Systems (ADAS) and Collision Avoidance Systems (CAS). Firstly, we introduce an enhanced inverse sensor model tailored for high-resolution laser scanners for building OGM. It compensates the unreflected beams and deals with the ray casting to grid cells accuracy and computational effort problems. Secondly, we introduce the 'vehicle on a circle for grid maps' map alignment algorithm that allows building more accurate local maps by avoiding the computationally expensive inaccurate operations of image sub-pixel shifting and rotation. The resulted grid map is more convenient for ADAS features than existing methods, as it allows using less memory sizes, and hence, results into a better real-time performance. Thirdly, we present an algorithm to detect what we call the 'in-sight edges'. These edges guarantee modeling the free space area with a single polygon of a fixed number of vertices regardless the driving situation and map complexity. The results from real world experiments show the effectiveness of our approach.**

## I. Introduction

The autonomous operation of a vehicle requires detailed knowledge about the environment. A versatile approach to this task is to use occupancy grid maps (OGM) introduced in [1]. Occupancy grid maps use a Bayes filter to estimate the occupancy probability for each grid cell to create a history effect that makes it robust against sensors uncertainty. There is a trade-off between the map resolution (accuracy) on one hand, and the memory and computational costs required for storing, updating, and retrieving the map on the other hand.

Bayesian grid maps are widely used to enable robust and consistent modeling of the local environment dynamic and static objects [2], [3]. Recent works have been using OGM to detect static environments [4], and to detect [3], classify [5], and track [2] dynamic objects. Most of the OGM-based works are using a laser scanner sensor [2], [3], [4], [5], while the framework can cope with other different sensor types like, as example, radars [6] and cameras using stereo vision [7]. The data used varies between real world data [2], [3], [4], [6] and synthetic data generated via computer simulators [5]. Grid maps rely on an inverse sensor model that maps sensor measurements back to its causes on the map. The biggest majority of the relevant works in literature rely on forward models [8] that deal with individual sensor measurements separately [2], [3], [4], [5], [6]. For intelligent vehicles applications with large map areas, the majority tends to rely on a local OGM and to fix the vehicle position within that map.

Drivable free space information is vital for autonomous vehicles and next generation Advanced Driver Assistance Systems (ADAS) and Collision Avoidance Systems (CAS) that have to plan evasive maneuvers in real-time [4]. To perform driving maneuvers autonomously, the vehicle has to efficiently achieve accurate free space detection online. In this paper, we propose a novel method for efficient extraction of static free space in a simple representation of a single polygon based on OGM. Relevant works in literature represented free space as Parametric Free Space Maps (PFS) [9], contours [4], a ray based free space [7], and pixels on camera images [10], [11].

Recently, vision-based ADAS have gained broad interest, thanks to the massive parallelization in modern GPU's, in combination with modern neural network architectures like Convolutional Neural Networks (CNN). Significant progress has been achieved in recent works in literature to detect drivable free space using a camera sensor as input [10], [11]. Such methods benefit from the rich visual information provided by the camera and requires data annotation. However, the detected free space range is limited to the driving scene captured by a single camera frame. Also, it's represented in the camera image coordinate space, and an additional level of complexity is required to transform representations back to world coordinate space. On the other hand, our proposed method uses a laser scanner sensor as input to detect the static free space surrounding the vehicle from all directions and in long ranges. The free space is provided directly in world coordinate space and doesn't require data annotation.

Firstly, we introduce an enhanced inverse sensor model tailored for high-resolution laser scanners for building OGM. Our inverse model copes with problems caused by beams overlapping and Moiré effect, and saves computational effort compared to the conventional inverse model. It works on raw laser full scans data and compensates the unreflected beams. Secondly, we introduce a novel algorithm ('vehicle on a circle for grid maps') that allows building more accurate local maps by avoiding the computationally expensive inaccurate operations of image sub-pixel shifting and rotation. The resulted grid map is more convenient for ADAS features than existing methods, as it utilizes higher proportion of the local map area for important vehicle direction based on its movement trajectory. Thirdly, we introduce an algorithm to detect what we call the 'in-sight edges'. These edges guarantee modeling the free space area with a single polygon regardless the driving situation and map complexity. A variant of the Douglas-Peucker line simplification algorithm [12] is used to fix the total number of vertices for the resulted polygon which is convenient to ADAS and automotive restrictions.

## II. ENHANCED INVERSE SENSOR MODEL FOR OGM

Although the basis of OGM is the independence of different grid cells, the measurement process can introduce correlations as a single observation may update several cells. The extent to which this happens is determined by the inverse sensor model in use. The biggest majority of the relevant works in literature rely on the conventional forward models [8] that deal with single sensor measurements separately. It is a common practice to use a simple ray casting model based on line drawing to determine the grid cells that are affected by a sensor measurement (a scan point) [13], [14], [15], [16]. A laser beam is characterized by its small angular perceptual field. Hence, in such common practice, a constant beam width which is set according to the previously chosen grid cell size is assumed. This is an incorrect assumption, since beams overlap multiple cells depending on the sensor characteristics and the cell-size of the grid.

Fig. 1 shows the overlapped cells of three adjacent laser beams in yellow color. The number of beams contributing to the probability of a single cell to be occupied is large in regions close to the sensor origin and vice versa as shown in Fig. 1(A). Especially that, in practice, the horizontal angle between two adjacent beams is much smaller than the example angle shown in Fig. 1 (typically 0.25° in the sensor we used for our experiments detailed in Section 5(A); Valeo ScaLa sensor). Also, as shown in Fig. 1(B), the overlaps cause some grid cells to be suggested as free by some beam and occupied by another one (a conflict). Moreover, in regions far from the sensor origin, beams overlap causes the undesired Moiré effect [17].

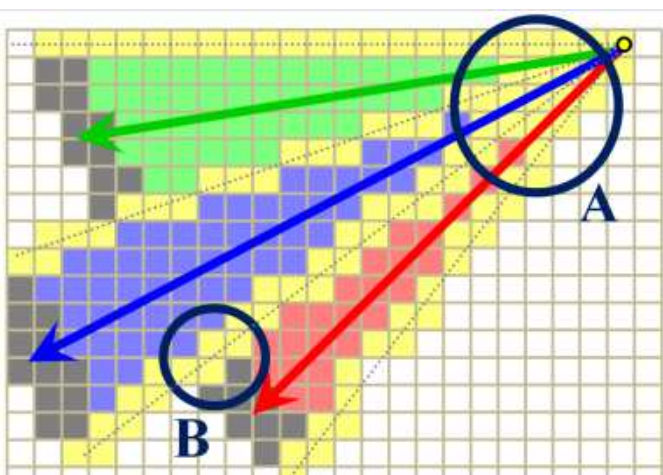

**Figure 1. Example for two OGM problems using an inverse sensor model based on individual measurements**

Raw laser scanner scan points are the union of scan points that stem from relevant targets and clutter. The former are vehicles, pedestrians, road boundaries, etc., whereas the latter are ground detections, rain and dust, and sensor noise. In order to create a robust algorithm for building up an occupancy grid map, it is paramount to remove the clutter before inserting it into the occupancy grid. For this purpose, we apply the DBSCAN (Density Based Spatial Clustering of Applications with Noise) algorithm [18] to the raw scan points to filter out clusters with a very small number of points (less than 5 points empirically). DBSCAN clusters scan points according to their density, it has the ability to determine the number of clusters in a data set. This makes it convenient for solving such clutter problem.

On the other hand, there are some expected sensor errors due to failures to detect any objects. This results into getting the maximum range readings from the sensor. In our inverse sensor model, these points are replaced by what we state as 'virtual' scan points. A virtual scan point is characterized by two characteristics. Firstly, its distance from the vehicle is equal to the minimum of distances from the vehicle to its nearest two non-virtual scan points before and after it. Secondly, unlike normal scan points, no obstacles are represented in the occupancy probabilities at the cells corresponding to such virtual points, while the area between the vehicle and those virtual scan points is updated as free space. Virtual scan points are shown in blue in Fig. 2(A).

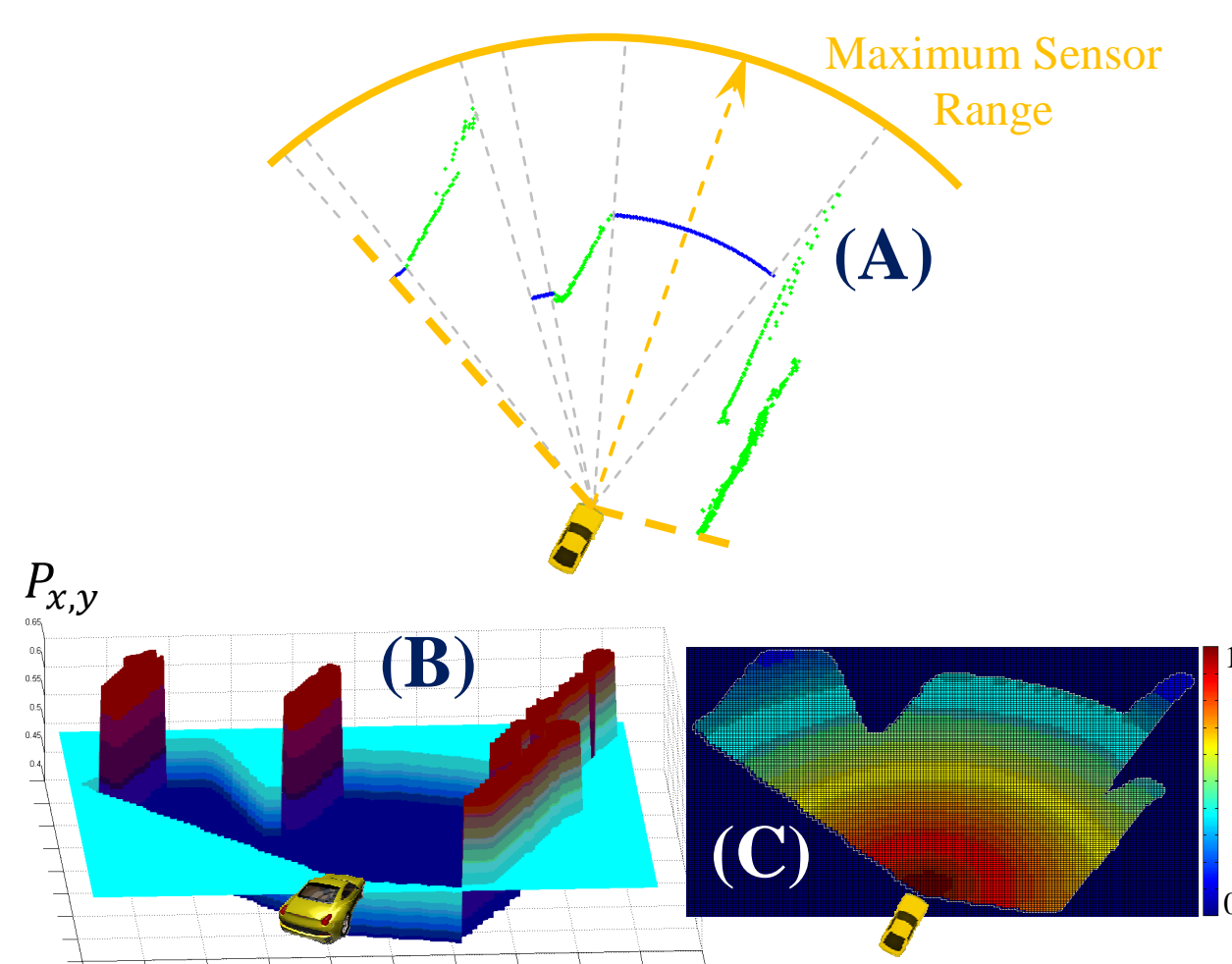

**Figure 2. Inverse sensor model based on full scans; (A) A full scan, where virtual scan points in blue, and normal scan points in green. (B) Occupancy Probabilities. (C) Occupancy Probabilities Confidences.**

In OGM, updating the map using estimated occupancy probabilities is done by a Bayes filter [1]. The grid cells area affected by any full scan is simply a polygon. Also curved boundaries, like the big blue curve of free space in the example in Fig. 2, are formed by small connected polygon edges. The Bresenham's line algorithm [19] is executed multiple times (at least three times when the area is exactly a triangle) to determine the cells to be updated due to some sensor reading. Having an inverse sensor model that handles a sensor full scan at once, saves computational effort caused by applying the Bresenham's line algorithm many times for each single scan point. Also, each map pixel is updated only once per full scan, instead of having many pixels updated more than one time in a single full scan because of beams overlap.

## III. VEHICLE ON A CIRCLE FOR GRID MAPS

Using a global grid map is convenient for robotics applications in a controlled area [20]. But for a high speed driving vehicle, it is not a convenient option due to limited memory and the irrelevance of old locations. Regardless of how big the memory storage is, eventually the vehicle will leave the map boundaries. If a map reset strategy is adopted then, 100% loss of map information occurs. In order to avoid a global map, a common approach in automotive mapping is to move the map with the host vehicle [21], [4]. However the notion of a moving map introduces some major computational burdens, namely rotation and shifting of the grids. Additionally, both operations lead to discretization errors and create artifacts that lower the overall map quality.

We avoid discretization errors due to fractional translations (sub-pixel shifting) by moving the host vehicle by the non-integer parts within the map, whereas errors due to rotation can be avoided by keeping the map orientation fixed to some global coordinate system and rotating the host vehicle.

The latter of these approaches requires a square map. Yet in high-velocity scenarios such as driving on a highway, ADAS functions are more interested in the environment in front of the car. In turn the square map reserves a lot of memory for regions of low interest. Our solution to this limitation is to grant even more freedom to the location of the host vehicle.

The key idea is to locate the sensor vehicle on a circle (shown in Fig. 3) with the vehicle orientation orthogonal to the circle tangent. The circle center itself is located in the center of the square map plus the fractional translation part. In the following, we will denote this circle 'vehicle position circle'. The size (radius) of the vehicle position circle and the angle on which the host vehicle is placed on it are determined by the velocity and the orientation of the vehicle respectively.

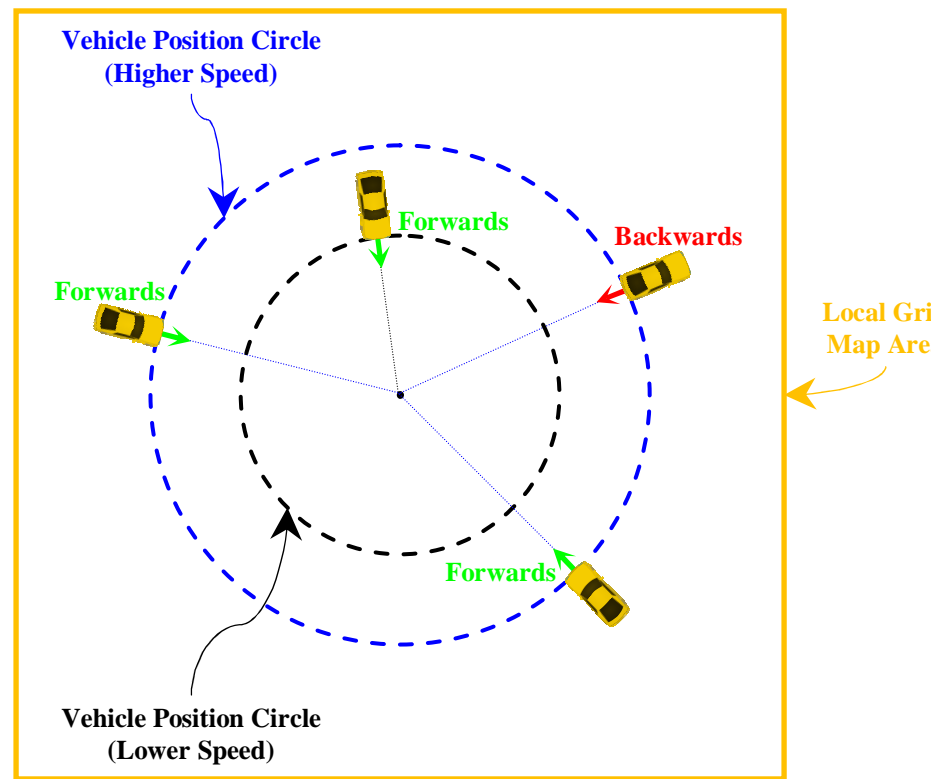

**Figure 3. Vehicle position circle**

With the position circle approach, we utilize the limited area covered in the local map to provide more information about the vehicle's area of interest. When the vehicle moves with higher speeds forward, the map is guaranteed to provide greater coverage in front of the vehicle than behind it. The amount of coverage is proportional to the vehicle speed. If the vehicle is moving backwards, the map provides greater coverage behind the vehicle than in front of it. This convention of vehicle position circle is more convenient for ADAS features than the conventional approaches of world coordinates and vehicle-centered maps. It dynamically allows the map to provide more coverage for the important area according to the vehicle velocity and direction.

The vehicle rotation (*yaw*) is the orientation angle of the vehicle relative an arbitrary starting orientation angle. Since the arbitrary starting position and angle are unknown, only the relative change of these coordinates matters. Using the odometry sensor (the detailed process is descried later in the Section 5(A) for system configuration), vehicle velocity magnitude (speed) is estimated. At each map update iteration $t$, the vehicle velocity $S_t$ is estimated using the change in the $x$ and $y$ coordinates of the vehicle position from the current to the previous update iteration. A low pass filter may be used to estimate the filtered vehicle velocity $\widehat{S_t}$ as in (1) to create a smooth trajectory.

$$\widehat{S_t} = \sum_{k=1}^{n} a_k . S_{t-k+1}, \sum_{k=1}^{n} a_k = 1. \tag{1}$$

For $\widehat{S_t}$ to be the mean value of the $t^{th}$ vehicle speed and its ($n$-$1$) preceding estimated speeds, $a_k = n^{-1}$. By comparing the vehicle position $x$ and $y$ coordinates at iteration $t$ with the vehicle position *yaw* angle of the preceding iteration, the movement direction (forward or backward) of the vehicle is obtained. $S_t$. It should be a negative value if the vehicle is moving backwards.

The equations modeling the local OGM and the vehicle position circle are as follows. At an iteration $t$, $M^t$ is the OGM, $P^t_{world}$ is the odometry information in world coordinates, $P^t_{local}$ is the vehicle position within the local map. The four inputs to our algorithm are: $P^t_{world}$ , $P^{t-1}_{world}$ , $P^{t-1}_{local}$ , and $M^{t-1}$. The two outputs are: $P^t_{local}$ and $M^t$.

As shown in Fig. 3, at each map update iteration $t$, the radius of the position circle is calculated to be proportional to $\widehat{S_t}$. According to the sign of $\widehat{S_t}$ and the *yaw* angle of the vehicle, the vehicle position on the position circle $P^t_c$ is determined. For map rotation, we avoid it by having a variable vehicle *yaw* within our non-rotating map:

$$P^t_{local}.yaw = P^t_{world}.yaw. \tag{2}$$

For map shifting, we avoid sub-pixel shifting by shifting the map with the integer part of the non-integer shift value:

$$P_{Shift} = P^t_{world} - P^{t-1}_{world} + P^{t-1}_{local} - P^t_c, \tag{3}$$

$$M^t = (L_H)^{\lfloor P_{Shift}.y \rfloor} . M^{t-1}, \tag{4}$$

$$M^t = [(U_W)^{\lfloor P_{Shift}.x \rfloor} . [M^t]^T]^T, \tag{5}$$

where equations (4) and (5) shifts the image $M^{t-1}$ down then left by the integer part of $P_{Shift}.y$ and $P_{Shift}.x$ respecivelly. $H$ and $W$ are the height and width of the local map $M$ in number of pixels respectively. $T$ denotes matrix transpose. $L$ and $U$ are the lower and upper shift matrices respectively.

Finally, vehicle position within that map is shifted by the fraction part of the non-integer shift value:

$$P^t_{local}.x = P^t_c.x + \{P_{Shift}.x\}, \tag{6}$$

$$P_{local}^{t}.y = P_{c}^{t}.y + \{P_{Shift}.y\}. \quad (7)$$

Initially: $P_{local}^{0} = P_{c}^{1}$, $P_{world}^{0} = P_{world}^{1}$.

Our method allows for an adaptive usage of the limited area of the local map, i.e. adjust the vehicle's area of interest according to the driving situation. When the vehicle moves with higher velocities forward, the map is guaranteed to provide greater coverage in front of the vehicle than behind it. It is important to stress that the amount of coverage is proportional to the vehicles velocity and hence matches typical use cases. Our method is, to our knowledge, the only consistent way to concurrently achieve the following:

1- avoiding the computationally expensive operations of image sub-pixel shifting and image rotation each map update cycle,
2- providing accurate (exact) local grid map with no image rotation nor sub-pixel shifting approximations,
3- ability to store smaller maps and save memory; as the big part of the map is guaranteed to cover the important information of the environment from ADAS point of view,
4- the world area covered in the map is mathematically proportional to its importance according to vehicle velocity and orientation,
5- and preventing map reset, because the vehicle will never leave the map boundaries despite moving around in it.

Fig. 4 shows two examples of the change of the position circle diameter at different map update iterations for two different vehicle trajectories. In each example, 40 map update iterations are recorded in different color for each iteration.

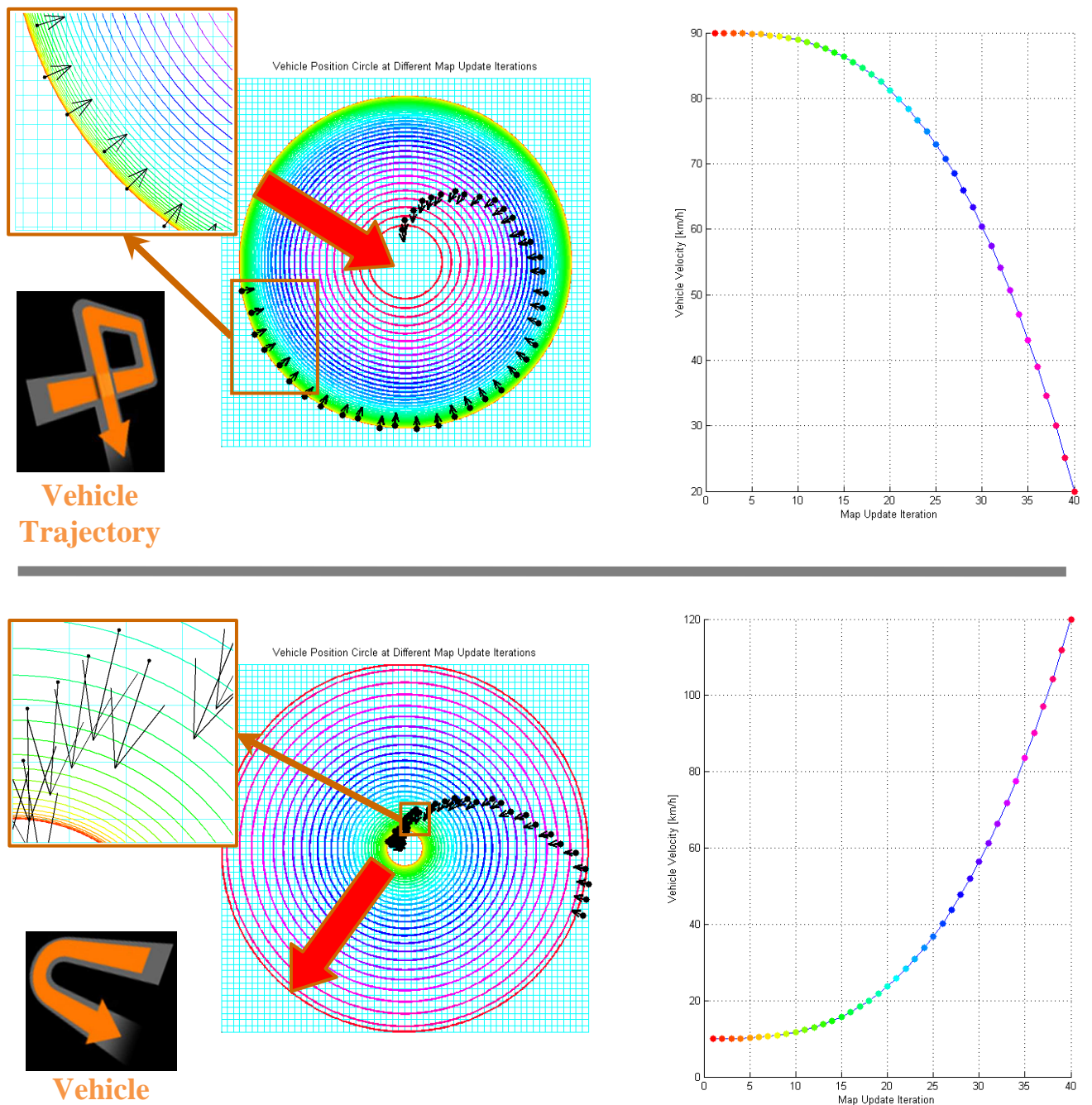

**Figure 4. Two examples of position circle at different map update iterations for different vehicle trajectories**

To conclude, the vehicle position circle algorithm can be viewed as a utility that can be used to position the vehicle within grid maps, regardless the algorithm used to build these maps. It's a pure map alignment method that saves computational time by avoiding image rotation and sub-pixel shifting. At the same time, it results into more accurate grid maps by avoiding approximations resulted from such two operations. A more accurate grid map leads to more accurate free space detections. On the other hand, the vehicle position circle algorithm allows adopting smaller grid map sizes, and hence, less memory consumption and better real-time performance. Because it utilizes higher proportion of the local map area for important vehicle direction based on its movement trajectory.

## IV. STATIC FREE SPACE EXTRACTION

### A. 'In-sight Edges' Detection

The possibility to represent arbitrary environments comes with the downside of high memory consumption which can create severe bottlenecks if the information has to be transmitted via an interface. In order to overcome this problem, we use a conversion from the grid-based environment representation to a feature-based representation. Such a representation has the additional advantage that it is preferred by most ADAS functions.

As in [1], a morphological opening image processing is performed on the OGM to separate small free areas that the vehicle cannot reach due to its dimensions. Then, what we call 'in-sight edges' detection is applied to the normalized occupancy grid map $\widehat{M^t}$ ($t$ is the map update iteration) as in Fig. 5; cells within $\widehat{M^t}$ are decided either free ('1'; foreground pixels) or occupied ('0'; background pixels) using neighborhood hysteresis thresholds. The figure shows three examples for the detected in-sight edges in different vehicle positions, where gray color represents free areas and white color represents occupied areas. The vehicle position within the local map is marked with red dot and its direction is shown in a yellow arrow, the position circle is shown with a dashed yellow circle, and the area of the in-sight edges is shown bounded in blue and filled with light gray.

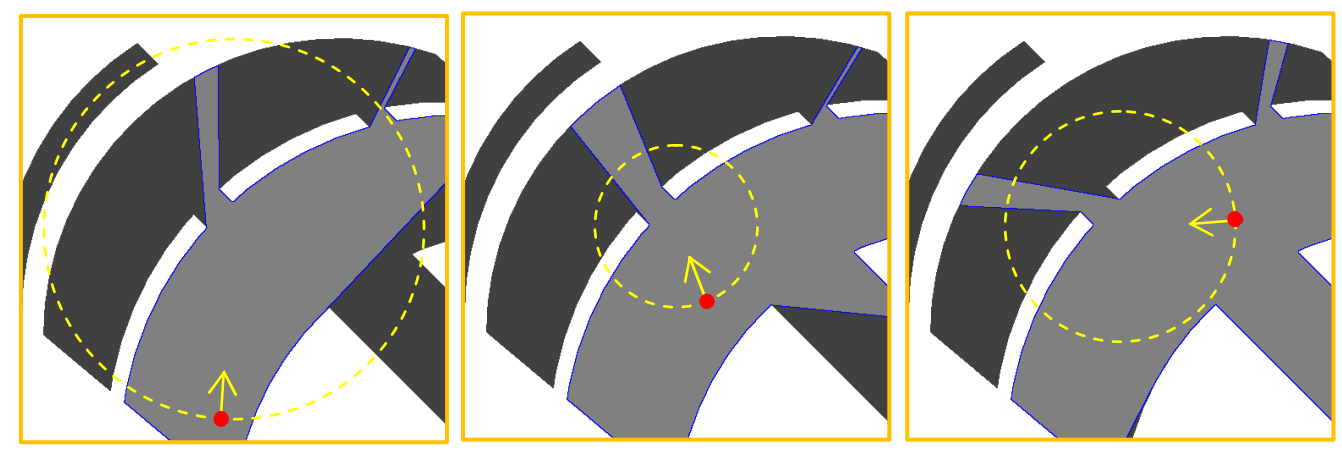

**Figure 5. In-sight edges Examples**

$C_{Edge}$ is the set of cells forming the in-sight edges, i.e., the pixels colored in blue in Fig. 5. The advantage of detecting such in-sight edges' cells is that it guarantees providing the free space area in a single polygon regardless the map complexity. If we sort the $C_{Edge}$ cells, the polygon connecting these sorted cells is a convenient and plain representation of free space for ADAS applications.

The Bresenham's line algorithm [19] can determine the grid map cells that should be selected in order to form a close approximation to a straight line between two points. The function *Bresenham*, used in the following algorithm, constructs the Bresenham's line starting from the start point towards the end point, and returns the coordinates (*y*,*x*) of the nearest occupied cell (pixel) it meets. If no occupied cells are met, it returns the end point. Given that $C_{\widehat{M}^t}^{i,j}$ is the cell at the $i^{th}$ row and the $j^{th}$ column of the map $\widehat{M}^t$, and that $H$ and $W$ are the height and width of that map in number of cells respectively, the algorithm for detecting the sorted in-sight edges' cells $C_{Edge}$ is as follows:

**algorithm** In-sight Sorted Edges Detection
**input:** Normalized map $\widehat{\boldsymbol{M}}^t$, Vehicle position $\boldsymbol{P}_{local}^t$
**output:** List of sorted in-sight edges' cells $\boldsymbol{C}_{Edge}$

**allocate** zero matrix $\boldsymbol{T}_{H\times W}$

**for** $\boldsymbol{i} := \mathbf{1}$ **to** $\boldsymbol{H}$ **do**
 **(y,x) = Bresenham( from** $\boldsymbol{P}_{local}^t$ **to** $\boldsymbol{C}_{\widehat{M}^t}^{i,1}$ **)**
 $\boldsymbol{T}^{y,x} \leftarrow \boldsymbol{i}$
 **(y,x) = Bresenham( from** $\boldsymbol{P}_{local}^t$ **to** $\boldsymbol{C}_{\widehat{M}^t}^{i,W}$ **)**
 $\boldsymbol{T}^{y,x} \leftarrow \mathbf{2}\boldsymbol{H} + \boldsymbol{W} - \boldsymbol{i} + \mathbf{1}$

**for** j:= **1 to** $\boldsymbol{W}$ **do**
 **(y,x) = Bresenham( from** $\boldsymbol{P}_{local}^t$ **to** $\boldsymbol{C}_{\widehat{M}^t}^{1,j}$ **)**
 $\boldsymbol{T}^{y,x} \leftarrow \mathbf{2}\boldsymbol{H} + \mathbf{2}\boldsymbol{W} - \boldsymbol{j} + \mathbf{1}$
 **(y,x) = Bresenham( from** $\boldsymbol{P}_{local}^t$ **to** $\boldsymbol{C}_{\widehat{M}^t}^{H,j}$ **)**
 $\boldsymbol{T}^{y,x} \leftarrow \boldsymbol{H} + \boldsymbol{j}$

$\boldsymbol{C}_{Edge}$ is the sorted set of coordinates of the non-zero pixels in **T**

**return** $\boldsymbol{C}_{Edge}$

Fig. 6 shows examples for $\widehat{M}^t$ and its $T$ matrix, and the resulting $C_{Edge}$ cells in $T$. The green dot in the figure represents the vehicle position within the local map $P_{local}^t$. The dashed arrow in Fig. 6(A) shows the direction of the sorted pixels forming $C_{Edge}$.

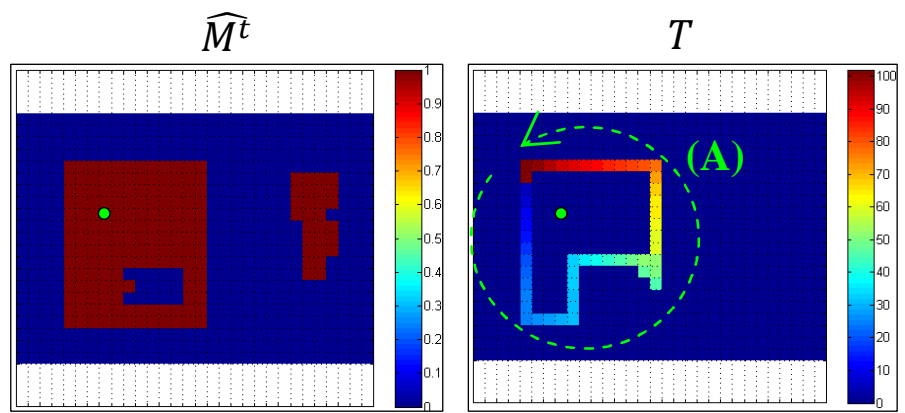

**Figure 6. In-sight edges detection algorithm**

## B. *Line Simplification*

The polygon formed by the coordinates of the in-sight edges sorted cells is a convenient representation for the static free space around the vehicle, where the center of each cell is a polygon vertex. To fix the number of vertices of that polygon, the Douglas-Peucker lines simplification algorithm [12] is used. The algorithm uses a point-to-line distance tolerance ε. It starts with a crude simplification that is the first and last vertices of original polygon. Then, for each two consecutive vertices in the simplification forming a line $L$, checks the intermediate vertices to select a vertex $V$ that makes maximum distance with $L$. If that distance is greater than ε, $V$ is added to the simplification. The process is repeated until all vertices of the original polygon are within a tolerance distance ε from the simplification.

From ADAS features and software points of view, it's convenient to define an upper limit for the number of vertices in the simplified polygon. We modify the original Douglas-Peucker algorithm to produce a maximum of $N$ number of vertices and to still use the tolerance parameter ε. The modification is solely adding another stop condition to the algorithm repetition; if the number of vertices in the simplification reaches $N$, the algorithm stops. Fig. 7 shows example results for the modified algorithm using different values for ε and $N$. The original polygon is shown in green color, and each image contains two results for simplification one in red color and the other one in blue color. In the examples, ε values are in meters, and the width and height of each image is 100 and 70 meters respectively.

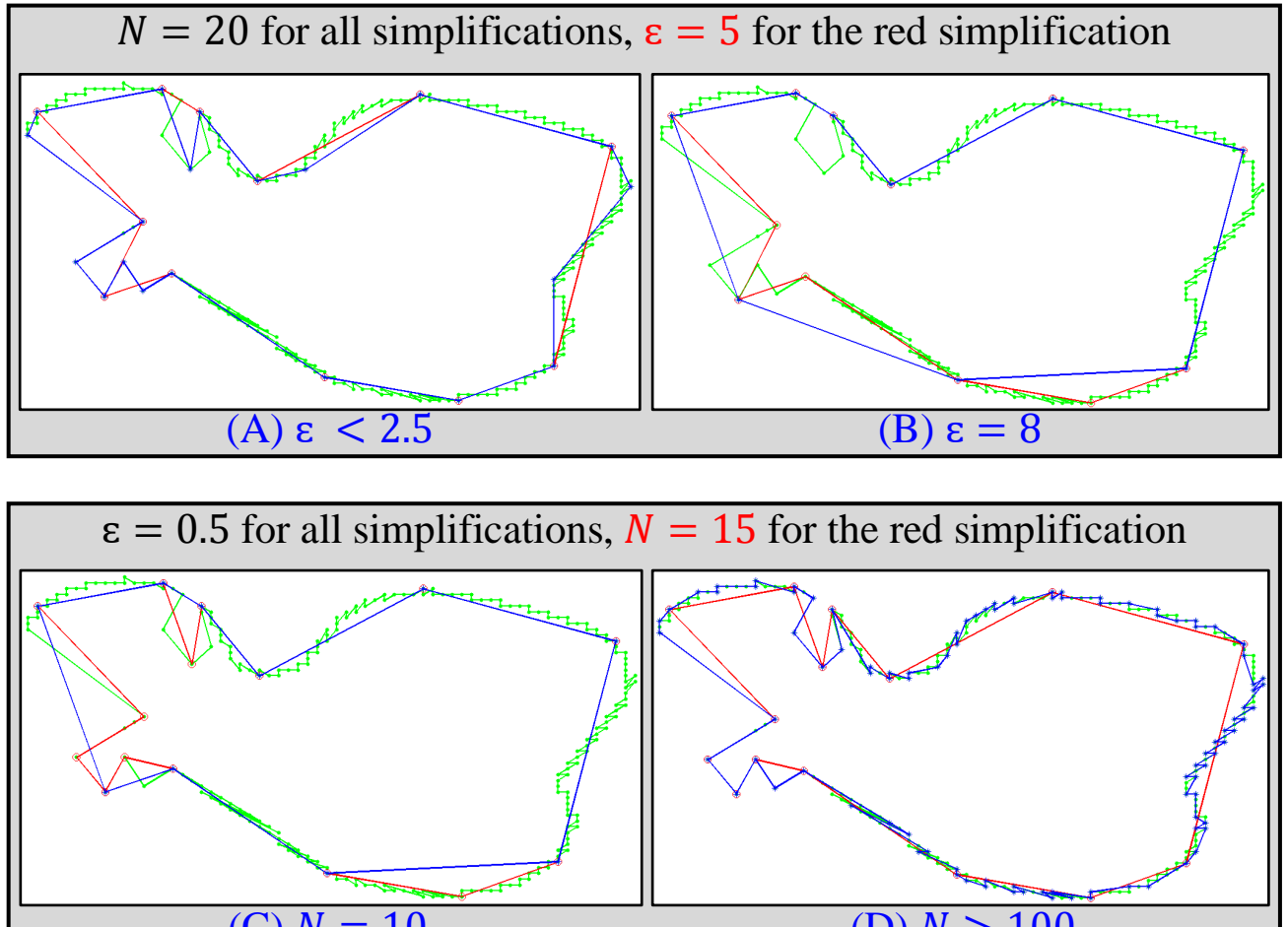

**Figure 7. The modified Douglas-Peucker algorithm**

The simplification results in Fig. 7(A) and (B) are using the same value for $N$ and different values of ε. As in Fig. 7(A), further decreasing ε produces the same polyline as the number of vertices $N$ vertices is already reached. The simplification results in Fig. 7(C) and (D) are using the same value for ε and different values of $N$. As in Fig. 7(D), further increasing $N$ produces the same polyline because the value of ε prevents accepting more vertices.

# V. EXPERIMENTAL RESULTS

## A. *System Configuration*

### *1) Laser Scanner*

Our test vehicle is equipped with a Valeo ScaLa which is a mass production automotive laser scanner. The sensor package size is 105x60x100 mm. It features four layers (planes) with 0.8° of separation and 145° of horizontal opening aperture with a horizontal resolution of 0.25°. It has multi-plane long range and embedded object tracking and classification to detect and track static environment as well as dynamic moving objects over a detection range of 150m. For this paper work, only the pure raw scan point data is used. Such points (as in Fig. 8) are primitive data that is provided by all the types of laser scanners regardless their technology, capabilities, and technical parameters.

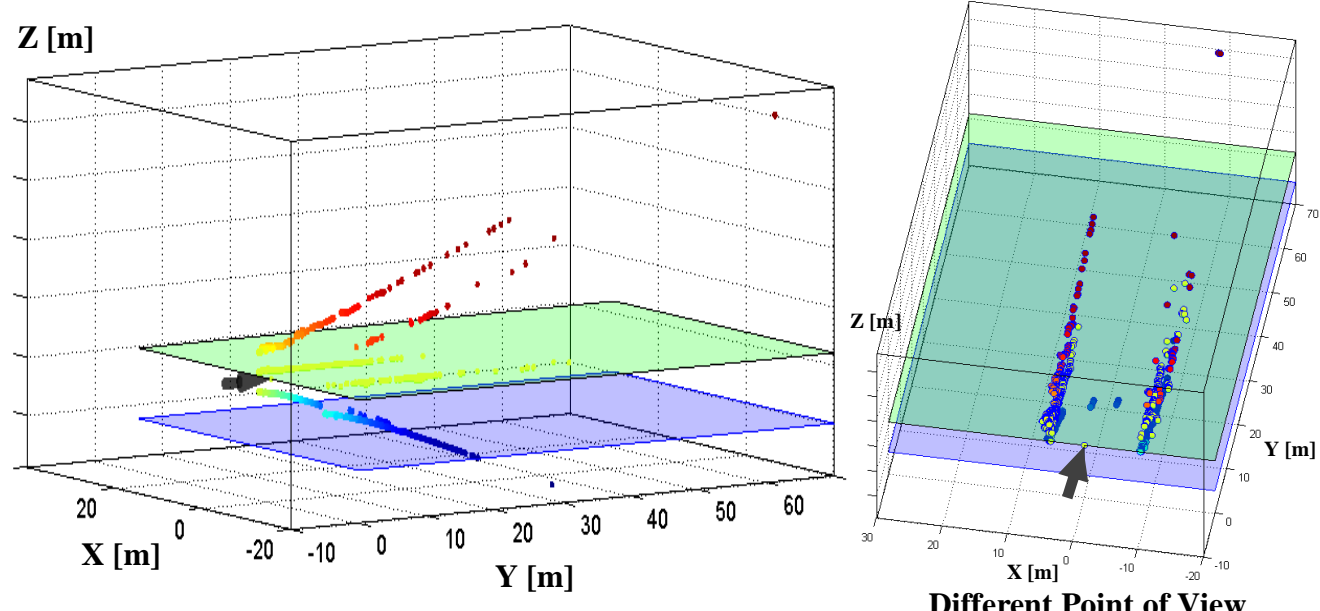

**Figure 8. A 3D visualization for a ScaLa sensor full scan**

Valeo ScaLa sensor uses a 905 nanometers-infrared laser that is safe for human eye. The sensor provides a full scan each 40 milliseconds and has a distance resolution less than 0.1 meters. The package size of the sensor is 105 x 60 x 100 millimeters. Fig. 8 shows an example for a ScaLa full scan. It's obtained from a real world scenario, where the sensor was mounted on vehicle's front at 0.3 meters height from physical ground. The dots in the figure represent the Cartesian coordinates of the scan points, where each scan point color is proportional to the dot z value; blue and red for lower and higher z values respectively. The sensor mounting position and orientation is shown with the back arrow. Each full scan, around 2000 scan points are measured representing 3 layers out of the sensor 4 layers.

*2) Odometry Sensor*

The second input to our system, is the vehicle odometry. Odometry represents the movement (position and orientation) of the host vehicle with respect to the world coordinate system. The vehicle position is represented by the $x$ and $y$ coordinates of the vehicle relative to an arbitrary starting point in the world coordinate system. The vehicle rotation (yaw) is the orientation angle of the vehicle relative an arbitrary starting orientation angle.

For our study, we use different vehicle sensors to calculate the odometry information. The odometry information within this approach is comprised of several sources; vehicle rotation (yaw) and brake pressure are estimated by the vehicle rotation rate sensor of the Electronic Stability Control (ESP) unit [22], the steering wheel angle is estimated by the steering angle sensor, and the vehicle velocity magnitude (speed) is the combined speed calculated from Transmission Control Unit. The odometry calculation from these inputs is based on a single track model with circular driving under static conditions. Empirically, the odometry accumulating errors over our 300x300 meters$^2$ local map are negligible. This is evident by the quality of the OGM maps shown in Fig. 9, 10, and 11.

## B. *Results*

To benchmark our enhanced inverse sensor model against the conventional one, each method is used to build an OGM based on real world test data, and then we compare their results. Our laser scanner is mounted front-facing on a Hyundai Grandeur car front bumper, and is used to collect data of 2 and 1 Kilometers of highway and city driving respectively, in Bietigheim in Germany. Our execution time analysis on this test data showed that our inverse sensor model is much faster; it processed full scans in less time than the conventional one by 44.07% on the average. Fig. 9 compares OGM snapshots from each method using same input data. The comparison shows that the enhanced inverse sensor model results into an OGM with less erroneous artifacts in most of the cases. For the conventional inverse sensor model results, areas highlighted with blue dotted eclipses highlight how not compensating unreflected beams is limiting the sensor vision. Red dotted eclipses highlight some OGM artifacts that reduce map quality, they are caused mainly because of beams overlap.

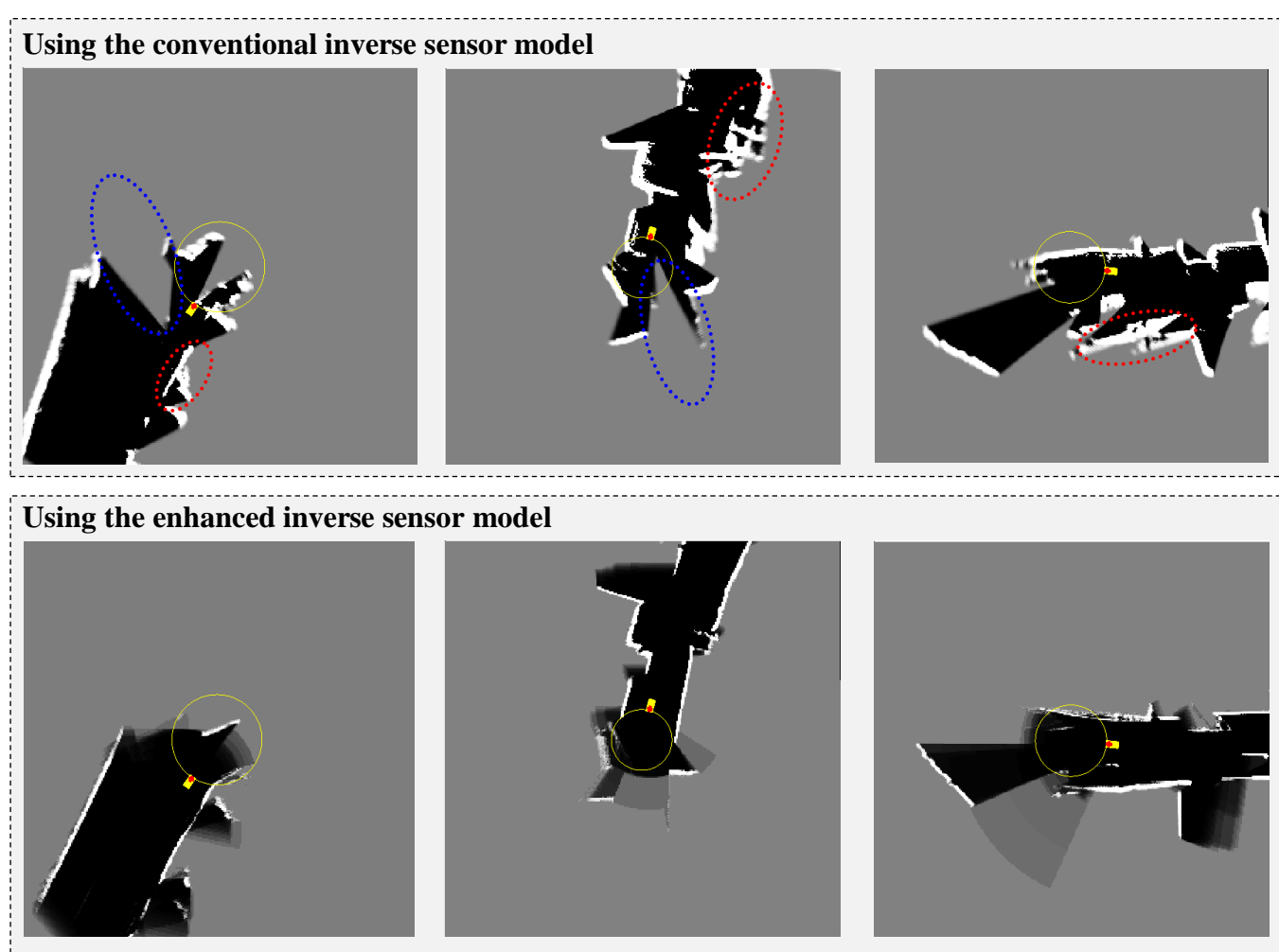

**Figure 9. OGM built using the enhanced inverse sensor model versus the conventional one. The enhanced inverse sensor model is proven much faster and the map quality is better.**

On the other hand, the overall time saved by the 'vehicle on a circle' algorithm by avoiding image rotations and sub-pixel shifting is equal to 4.30% of the overall OGM execution time (using the enhanced inverse sensor model). This save percentage increases if the used image rotation and sub-pixel shifting algorithms are more sophisticated. Fig. 10 shows the OGM resulted by our enhanced inverse sensor model along with the 'vehicle on a circle' in yellow. The figure shows two examples out of hours of testing. The laser scanner data for both of the two examples are recorded in Bietigheim in Germany. The first one is for urban city driving and the second one is for high-way driving. The used occupancy grid has a size of 300 × 300 cells. The cell resolution was set to 0.2 meters. The inverse sensor model probabilities of free, 'no information', and occupied are 0.40, 0.5, and 0.65 respectively.

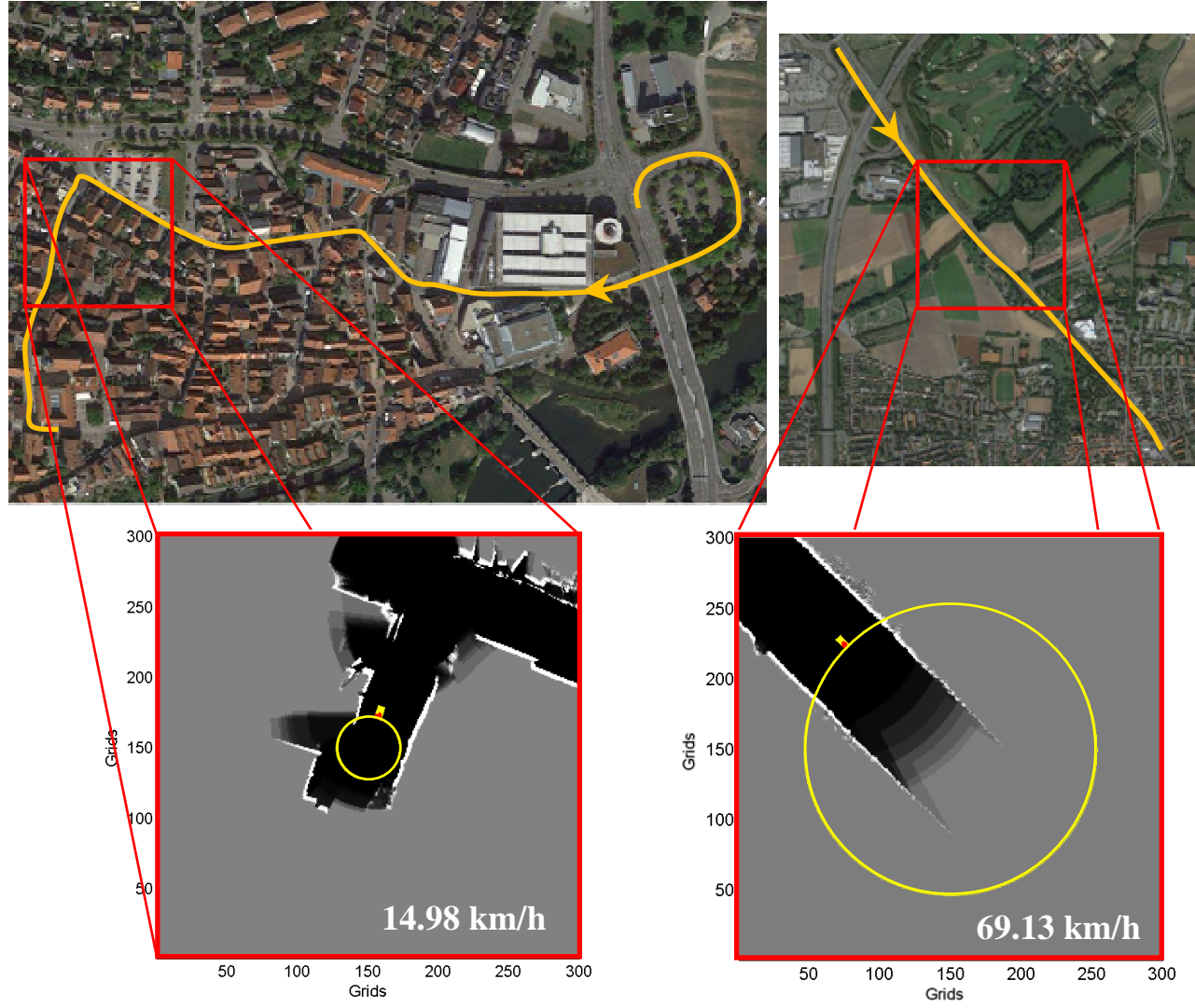

**Figure10. Real world OGM results on an urban and a highway driving scenarios with the vehicle position circle shown in yellow**

Fig. 11 shows three snapshots for the extracted free space polygon at different vehicle speeds. The free space polygon is overlaid with vehicle north directed real satellite map. The examples in Fig. 11 (A) and (B) are for areas that are decided as free by the map while it's not included in the free space polygon. This is a limitation of the in-sight edges detection algorithm detailed in section 4(A) to allow the free space to be modeled in a single polygon in any complex driving scenario. In Fig. 11(C), there is real free space area that is not represented in the map. This is due to the fact that in that direction laser beams are unreflected, and when compensated as virtual scan points by our inverse sensor model, the distance from the vehicle to a virtual scan point is equal to the minimum of distances from the vehicle to its nearest two non-virtual scan points before and after it (as detailed in section 2). This is limiting the sensor range in some conditions like that example, but in general, our experiments prove that this method allows for more accurate and confident maps.

## VI. Conclusion

We introduce a new efficient method for environmental free space detection with laser scanner to be used for Advanced Driving Assistance Systems (ADAS) and Collision Avoidance Systems (CAS). Firstly, our method is based on 2D occupancy grid maps (OGM) built using an enhanced inverse sensor model that is tailored for high-resolution laser scanners and works on preprocessed raw laser scanner data. It resulted into more accurate maps and saved computational effort. Secondly, we introduce the 'vehicle on a circle for grid maps', which is a pure map alignment algorithm, that allows building more accurate local maps while saving computational effort. The resulted map also is more convenient for ADAS features than existing methods, as it utilizes higher proportion of the local map area for important vehicle direction based on its movement trajectory. Finally, we introduce an algorithm to detect what we call the 'in-sight edges' to guarantee representing the free space in a single polygon of a fixed number of vertices regardless the driving situation and map complexity.

The results from real world experiments show the effectiveness of our method, it allows detecting the drivable free space in a simple format in real-time with high accuracy which is vital for autonomous driving vehicles that have to plan evasive maneuvers.

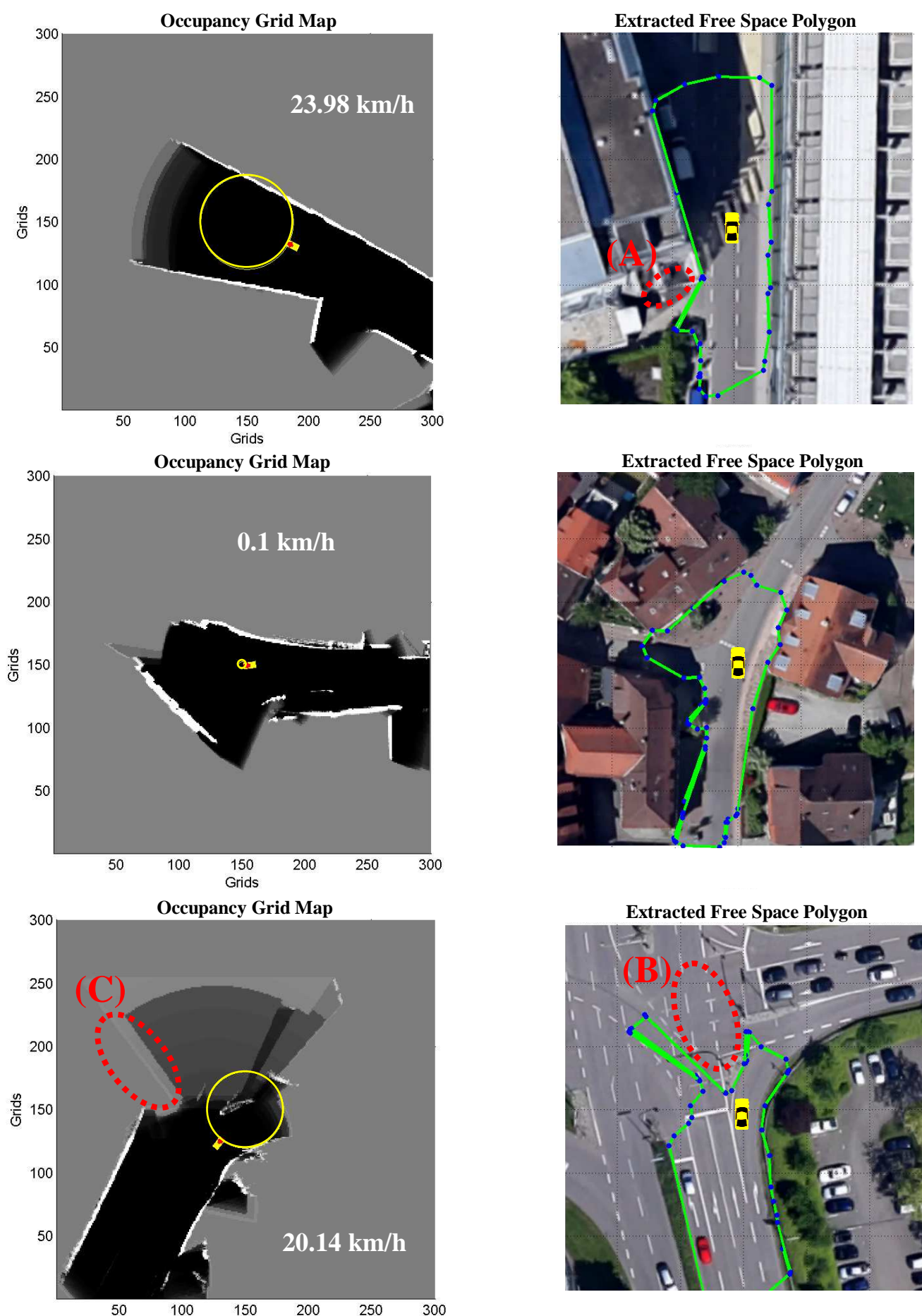

**Figure 11. Detected free space at different vehicle speeds overplayed with satellite maps**